\documentclass[letterpaper]{article} 
\ifdefined\aaaianonymous
    \usepackage[submission]{aaai2026}  
\else
    \usepackage{aaai2026}              
\fi
\usepackage{times}  
\usepackage{helvet}  
\usepackage{courier}  
\usepackage[hyphens]{url}  
\usepackage{graphicx} 
\usepackage{natbib}  
\usepackage{caption} 
\usepackage{algorithm}
\usepackage{algorithmic}
\usepackage{amsmath}
\usepackage{booktabs}
\usepackage{newfloat}
\usepackage{listings}
\DeclareCaptionStyle{ruled}{labelfont=normalfont,labelsep=colon,strut=off} 
\floatstyle{ruled}
\newfloat{listing}{tb}{lst}{}
\floatname{listing}{Listing}
\title{Gentle Manipulation Policy Learning via Demonstrations from VLM Planned Atomic Skills}
\author{
    Jiayu Zhou\textsuperscript{\rm 1}\equalcontrib,
    Qiwei Wu\textsuperscript{\rm 2}\equalcontrib,
    Jian Li\textsuperscript{\rm 2},
    Zhe Chen\textsuperscript{\rm 1},
    Xiaogang Xiong\textsuperscript{\rm 1}\corr,
    Renjing Xu\textsuperscript{\rm 2}\corr
    \\
}
\affiliations{
    \textsuperscript{\rm 1}Harbin Institute of Technology (Shenzhen), Shenzhen, China.\\
    \textsuperscript{\rm 2}Hong Kong University of Science and Technology (Guangzhou), Guangzhou, China. \\
}

\usepackage{bibentry}

\begin{document}

\maketitle

\begin{abstract}
Autonomous execution of long-horizon, contact-rich manipulation tasks traditionally requires extensive real-world data and expert engineering, posing significant cost and scalability challenges. This paper proposes a novel framework integrating hierarchical semantic decomposition, reinforcement learning (RL), visual language models (VLMs), and knowledge distillation to overcome these limitations. Complex tasks are decomposed into atomic skills, with RL-trained policies for each primitive exclusively in simulation. Crucially, our RL formulation incorporates explicit force constraints to prevent object damage during delicate interactions. VLMs perform high-level task decomposition and skill planning, generating diverse expert demonstrations. These are distilled into a unified policy via Visual-Tactile Diffusion Policy for end-to-end execution. We conduct comprehensive ablation studies exploring different VLM-based task planners to identify optimal demonstration generation pipelines, and systematically compare imitation learning algorithms for skill distillation. Extensive simulation experiments and physical deployment validate that our approach achieves policy learning for long-horizon manipulation without costly human demonstrations, while the VLM-guided atomic skill framework enables scalable generalization to diverse tasks.
\end{abstract}

\section{Introduction}
Robotic manipulation has made significant progress, yet increasingly complex tasks demand both long-horizon continuous operation and fine-grained, contact-rich interactions. Traditional methods primarily rely on coarse visual perception, which falls short in capturing the delicate contact dynamics essential for precise manipulation \cite{DBLP:ChiFDXCBS23,Ze20243DDP}. Tactile sensing offers high-resolution surface and contact state information critical for enabling gentle and accurate physical interactions \cite{Xue2025Reactive, huang20243dvitac}. As manipulation tasks evolve from short, discrete stages to extended sequences requiring sustained dexterity and adaptability \cite{shi2023robocook}, addressing temporal dependencies and ensuring safe, gentle contact over prolonged periods becomes increasingly challenging.

Currently, long-horizon gentle manipulation remains a challenge. In imitation learning-based methods, the acquisition of expert demonstration data remains a major bottleneck due to its high cost and labor intensity. While visual-language model frameworks have shown promise in automating task planning and policy learning, current approaches largely overlook the integration of tactile and contact force information. The difficulties in achieving manipulation tasks through these methods can be summarized as follows. First, most existing public data benchmarks lack consideration and focus on force, yet in practical tasks, it is necessary to avoid damaging objects during manipulation. Second, perceptual uncertainty arising from sensor noise and partial observability complicates reliable decision-making. Finally, current data acquisition methods are expensive, and there are no effective ways to synthesize and expand data.

\begin{figure*}[htp]
    \centering
    \includegraphics[width=0.8\linewidth]{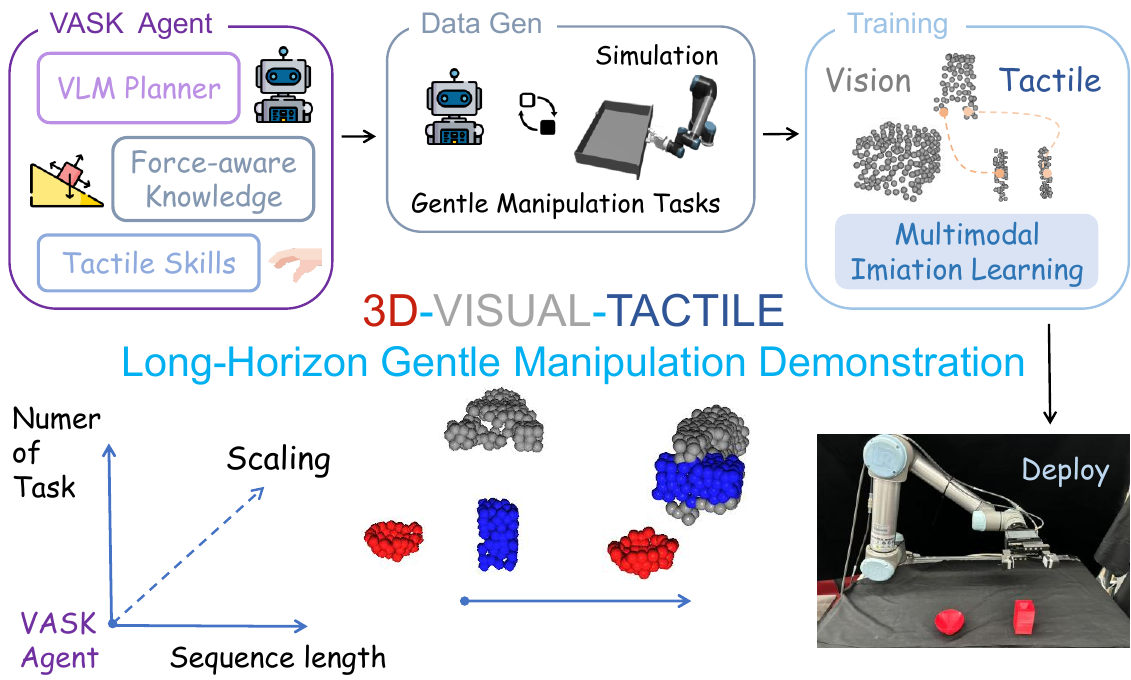}
    \caption{Our pipeline trains robotic arm tactile skills through force-constrained reinforcement learning in simulation. Visual Language Models then plan task sequences by interpreting visual scenes and language instructions to generate expert demonstrations. These demonstrations are distilled into contact-aware manipulation policies via visual-tactile Diffusion Policy, enabling end-to-end execution of long-horizon tasks from multi-modal point cloud inputs.
    }
    \label{fig:Pipeine}
\end{figure*}

 In this work, we propose a force-aware synthetic data generation approach through simulation-trained tactile atomic skills, VLM-guided hierarchical task decomposition, and the generation of multimodal long-horizon manipulation data, establishing a benchmark methodology for imitation learning to address these challenges, as shown in Fig.~\ref{fig:Pipeine}. Our main contributions include:
\begin{itemize}
    \item \textbf{Force-aware and scalable data generation:} We present a long-horizon sequential manipulation data generation framework that explicitly incorporates contact force information, improving reliability and safety while substantially reducing manual effort in constructing expert demonstrations. Leveraging a library of robust atomic skills, the framework supports scalable extension to diverse complex tasks through hierarchical composition.
    \item \textbf{Dataset and Training Benchmark VT-DP:} Based on our data generation method, we have prepared a small, scalable dataset with language labels that is based on visual-tactile 3D point clouds, and provided a benchmark VT-DP (Visual-Tactile Diffusion Policy) capable of enabling multi-modal imitation learning training. 
\end{itemize}
We conducted ablation studies to demonstrate the functionality of each component in our data synthesis framework. Additionally, we presented comparisons with different data collection methods, comparisons with different training frameworks, and successful policy deployment in the real world. Experimental results demonstrate that the final policy of our framework achieves superior performance, robust generalization capability, and significant extensibility.

\section{Related Work}
\subsubsection{{Contact-rich Long-Horizon Manipulation:}}
In robotic manipulation, traditional control methods typically require accurate dynamic models to perform complex tasks \cite{Li2024DynamicPF, Jin2022TaskDrivenHM, Jiang2024ContactImplicitMP}. However, obtaining precise models is often infeasible in real-world scenarios due to model uncertainty and environmental variability. Reinforcement learning algorithms \cite{Liu2023TactileAI,Chebotar2023QTransformerSO} have demonstrated the ability to learn manipulation policies through repeated interaction with the environment without relying on explicit models. Despite these advances, most RL research has focused on single-stage tasks such as grasping and pushing \cite{Zhong2025DexGraspAT,10182274,WuPZSXL24,10260563}. For contact-rich tasks, studies often emphasize peg-in-hole insertion and in-hand manipulation \cite{Ankile2024FromIT,Geng2023UniDexGraspID}, which remain relatively constrained problems.
As manipulation tasks become more complex and long-horizon in nature, designing appropriate reward functions becomes increasingly challenging, making it difficult for RL agents to converge to stable and effective policies. To address this, imitation learning \cite{Xue2025ReactiveDP,DBLP:ChiFDXCBS23,Zhao2023LearningFB,Ze20243DDP} has been widely adopted for long-horizon tasks by leveraging expert demonstrations to guide the agent. Nevertheless, imitation learning approaches typically require extensive and costly data collection from manual demonstrations, limiting scalability.
Unlike prior works that rely exclusively on reinforcement or imitation learning, our framework leverages the strengths of both. We first train robust single-stage tactile atomic skills via reinforcement learning in simulation. Complex long-horizon tasks are then decomposed into sequences of these primitives, enabling the generation of high-quality expert demonstration data. Finally, we distill these demonstrations through imitation learning into a stable, gentle manipulation policy capable of long-horizon execution. This hybrid approach reduces data collection and training burdens while improving adaptability and performance in challenging, contact-intensive long-horizon manipulation tasks.

\subsubsection{Visual Language Models for Data Generation:}
Integrating VLMs into robotic systems \cite{Chen2025Robo2VLMVQ,Yang2025LoHoVLAAU,Kim2024OpenVLAAO} significantly enhances intelligence and generalization by translating natural language into executable action sequences. Recent approaches leverage foundation models to automatically generate diverse tasks and scenarios, scaling robot datasets with minimal human input. RoboGen \cite{Wang2023RoboGenTU} utilizes language descriptions to generate scene configurations and learns skills through LLM-synthesized reward functions. GenSim \cite{Wang2023GenSimGR} employs LLMs for procedural generation of scenes, simulations, and demonstrations, while GenSim2 \cite{Hua2024GenSim2SR} extends this capability to long-horizon articulated tasks through enhanced reasoning models. HumanoidGen \cite{jing2025humanoidgen} further applies this paradigm to bimanual dexterous manipulation using pre-trained sub-policies and annotated 3D assets. However, these existing methods generally do not incorporate contact force information. In contrast, our framework explicitly integrates contact force data by leveraging pretrained atomic policies trained with force-aware reward functions combined with VLM-based compositional planning. This novel integration not only facilitates stable and physically consistent low-level execution but also enables the automatic generation of high-quality demonstrations for complex long-horizon manipulation tasks involving rich physical interactions.

\section{Method}
\subsection{Skills for Gentle manipulation}

\subsubsection{Contact Force Characterization}
Contact force is fundamental to robotic manipulation, especially in tasks requiring fine-grained interaction. Typically, contact force can be decomposed into three primary components: normal force, shear force, and torque, each exhibiting distinct characteristics during different manipulation phases. During grasping, the focus lies on the normal force generated by tactile sensors compressing the object's surface.  When the gripper moves the object, the resulting contact forces comprise a combination of normal force, shear force, and torque due to the multidirectional nature of the movement. This classification facilitates targeted analysis and optimization of the robot-object interaction for specific tasks, thereby enhancing both safety and operational efficiency.
\subsubsection{Force-aware Atomic Skills}
\begin{figure}[htp]
    \centering
    \includegraphics[width=1.0\linewidth]{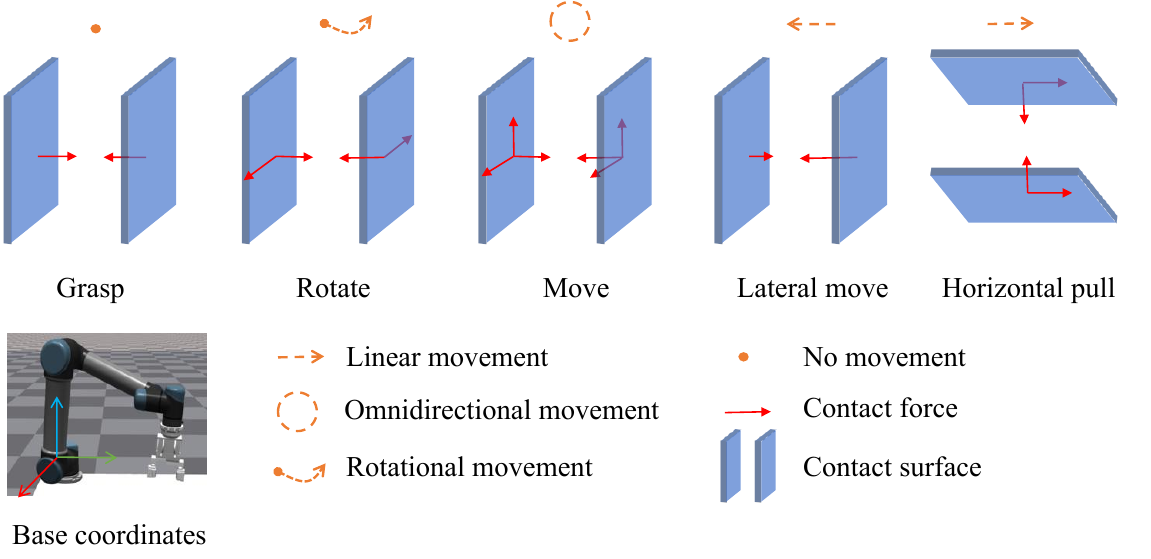}
    \caption{Schematic diagram of contact force of tactile atomic skills.}
    \label{fig:skills}
\end{figure}

As shown in Fig.~\ref{fig:skills}, leveraging the categorization of contact forces, we designed atomic skills that align with interaction patterns such as ``grasp", ``rotate", and ``move", with each skill focusing on specific force components. To accomplish more complex manipulation tasks, such as opening drawers and boxes, we developed the skill ``Horizontal pull", which primarily involves clamping force and horizontal friction, and the skill ``Lateral move", which combines vertical grasping force with unidirectional pushing force. Together, these atomic skills encompass the diverse contact force distributions encountered in real-world operations, thereby enabling robust task execution.

\subsubsection{Implementation of Atomic Skills}
Reinforcement learning (RL) provides significant advantages for acquiring atomic skills in contact-rich manipulation: its trial-and-error paradigm enables autonomous policy optimization through environmental interaction while accommodating stochastic task dynamics. To leverage these benefits, we implement atomic skill training using the Soft Actor-Critic (SAC) algorithm \cite{pmlr-v80-haarnoja18b}. SAC incorporates the maximum entropy principle to encourage policy stochasticity—critical for robust contact interactions—while its twin Q-network architecture reduces value estimation bias to enhance training stability. To further improve policy robustness, we apply domain randomization by dynamically perturbing key physical parameters during training, including object pose initialization, mass, and friction coefficients.

All training is performed within the Isaac Gym simulation environment under full state observability. The observation space includes the robot state, object state data, and tactile sensor contact forces. The action space consists of relative Cartesian commands to the end effector, denoted as $ \{ {a_x},{a_y},{a_z},{a_{rx}},{a_{ry}},{a_{rz}},a_c\} $, where translational ($a_x$, $a_y$, $a_z$) and rotational ($a_{rx}$, $a_{ry}$, $a_{rz}$) components operate along the Tool Center Point (TCP) axes, and $a_c$ controls the relative position of the width of the parallel gripper.
Each atomic skill is associated with a specifically designed reward function tailored to its single-stage objective, which incorporates a force penalty term to encourage gentle manipulation policies. For example, the reward function for the grasping skill is formulated as follows:
\begin{equation}
\begin{split}
R_{Grasp} &= T - \tanh(D_{e}) +PC_{p}\\
&+H_r + C_{r} + F_{p} + S_{r}
\end{split}
\end{equation}
where $T$ denotes a time penalty to encourage task efficiency, and the hyperbolic tangent function $\tanh(\cdot)$ normalizes the distance $D_{e}$ between the gripper fingers and the target object to ensure smooth reward variation. $PC_{p}$ penalizes premature gripper closure, while $C_{r}$ incentivizes maintaining contact with the object. $H_r$ rewards the height of the lifted object, $F_{p}$ penalizes excessive contact force, and $S_{r}$ provides a large positive reward upon successful task completion.
\subsection{Generation of Gentle Manipulation Data}
\subsubsection{Planning with Atomic Skills}
Constructing demonstration datasets for long-horizon manipulation tasks typically requires substantial manual effort and incurs high costs. To address this challenge, 
we propose a novel framework, Visual Language with Atomic Skills (VASK), which employs a pre-trained VLM as the top-level planner. We design a comprehensive knowledge base for the VLM that includes an operational context introduction, an atomic skill library, and system prompts. The atomic skill library encompasses various tactile skills, along with detailed descriptions of each skill and their corresponding contact force information during manipulation. For example, the skill ``Grasp" involves ``\textit{Move the gripper to the object and perform a grasp, considering normal contact forces to avoid damaging the object."}
During the planning phase, based on a human-provided task description and operational background image, VASK decomposes the task into corresponding sequences of atomic skills. For example, given the task description: \textit{``Open the drawer and place the object inside”}, the VLM first analyzes the drawer-opening operation: \textit{``This operation involves grasping the drawer handle and pulling it horizontally”}, Then it reasons: \textit{``The skill Horizontal pull, which considers normal contact force and lateral friction, precisely matches this scenario”}. Therefore, the first selected action is ``Horizontal Pull". Subsequently, according to the following task requirements, the model sequentially selects appropriate tactile skills, resulting in an action sequence such as [``Horizontal Pull”, ``Grasp”, ``Move”, ``Open”]. VLM explicitly accounts for the types of contact forces involved in each subtask and searches the atomic skill library to find corresponding tactile skills, thereby planning a reasonable and effective manipulation strategy.

\subsubsection{VASK for Data Generation}
\begin{figure}[htp]
    \centering
    \includegraphics[width=1.0\linewidth]{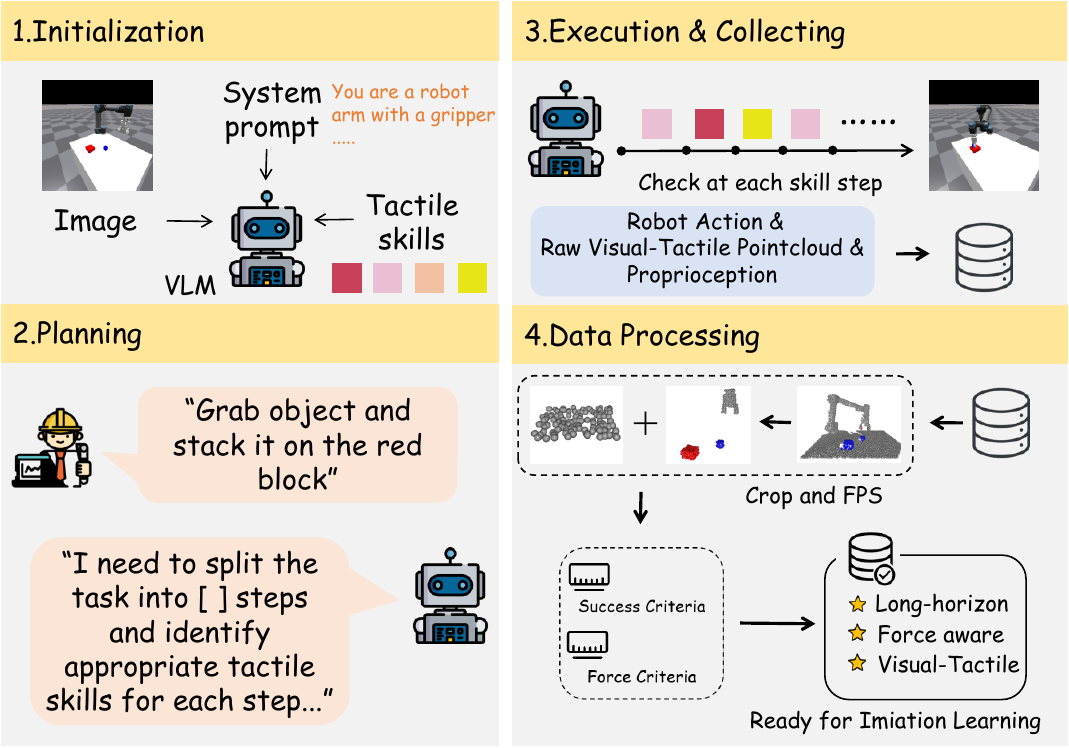}
    \caption{Framework diagram of VASK. The VLM receives system prompts, natural language task descriptions, and RGB images, and integrates this information with atomic skills to guide the agent in completing manipulation tasks and collecting raw trajectory data.}
    \label{fig:VASK}
\end{figure}

As illustrated in Fig.~\ref{fig:VASK}, the complete data generation process comprises four stages: \textbf{\textit{Initialization:}} The manipulation environment is initialized, and the VLM reads information from the knowledge base to guide hierarchical decomposition. \textbf{\textit{Planning:}} Generating hierarchical skill sequences via VLM by integrating natural language instructions and background image. \textbf{\textit{Execution and Collecting:}} The robot sequentially executes atomic skills while continuously acquiring environmental information and contact force feedback, simultaneously recording raw point cloud data during the manipulation process.
\textbf{\textit{Data Processing:}} Select trajectories that meet the task completion criteria and exhibit relatively low contact forces to ensure high-quality data and promote gentle manipulation.
It is worth noting that our approach uniquely incorporates contact force information into the long-horizon task data generation framework while significantly reducing the manual cost required to construct expert demonstration datasets.
\subsection{Learning Long-horizon Gentle Manipulation}
The VASK framework is capable of completing long-horizon gentle manipulation tasks, largely benefiting from full state observability available in the simulation environment. However, as task complexity increases, the inherent hallucination tendencies of the VLM lead to instability in the policy outputs. To address this, we employ knowledge distillation to extract operational policies grounded on visuo-tactile point clouds and proprioceptive sensing.
\subsubsection{Fusion of Visual and Tactile Modalities}
The fusion of visual and tactile modalities enables robotic systems to integrate global operational awareness with fine-grained contact details, facilitating complex manipulation tasks. Compared to RGB images providing limited 2D information, point clouds offer robust spatial geometric representations exhibiting strong invariance to illumination variations, background clutter, and color changes. In simulation, visual point clouds are acquired from depth images captured by fixed-mount depth cameras. For tactile point clouds, we adopt the soft contact model from TacSL \cite{Akinola2024TacSLAL} to simulate contact interactions. In this model, objects are still modeled as rigid bodies but are allowed to mutually penetrate in proportion to the interaction forces. Depth cameras capture these penetration states between rigid bodies to construct tactile point clouds that represent the contact geometry. Tactile point clouds deform in response to contact.
\subsubsection{VT-DP for Knowledge Distillation}
Diffusion models offer substantial advantages for imitation learning in robotic control, as they effectively capture complex, multimodal action distributions and ensure stable training through a gradual denoising process. Prior work DP3 \cite{Ze20243DDP} demonstrates that leveraging point clouds significantly enhances policy generalization in unstructured environments, making them particularly suitable for diffusion-based policy learning.
Building upon these insights, we propose Visual-Tactile Diffusion Policy (VT-DP), a long-horizon gentle manipulation distillation framework that integrates visual and tactile point clouds. Building upon the DP3 method, We crop the global point clouds to retain only the end-effector gripper and the manipulated object, and downsample the visual point cloud to 256 points as well as the tactile point cloud to 128 points. Furthermore, visual and tactile point clouds are encoded separately to better capture modality-specific features. Based on the VT-DP framework, only 50 high-quality demonstration trajectories are sufficient to distill gentle manipulation policies for corresponding long-horizon sequential tasks.
\begin{figure*}[htp]
    \centering
    \includegraphics[width=0.95\linewidth]{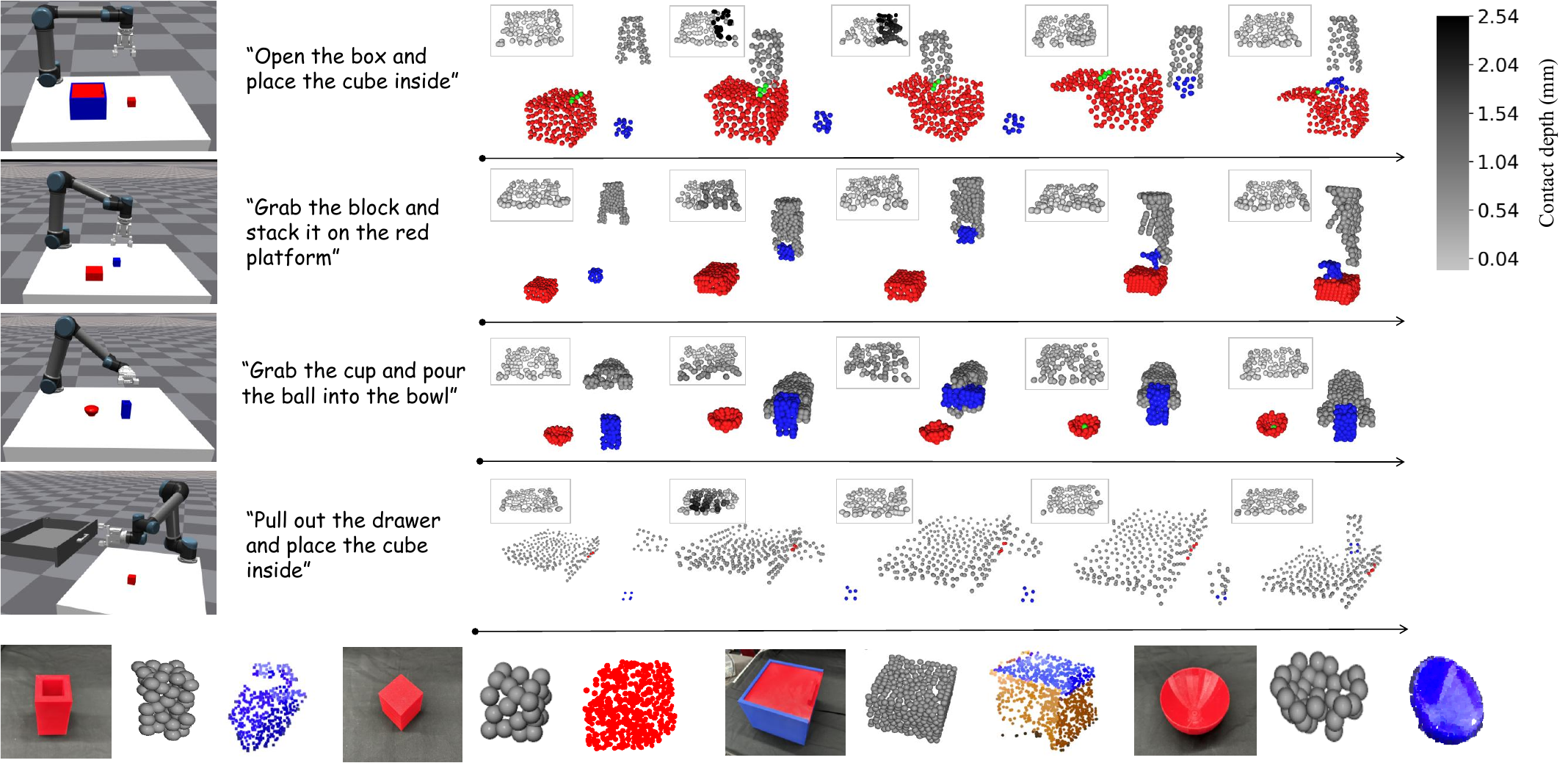}
    \caption{Point cloud sequence during a long-horizon manipulation task. The boxed region in the upper-left corner of the visual point cloud shows the tactile point cloud. The legend indicates the contact depth between the sensor and the object. Note that the point cloud data itself is uncolored; colors are applied solely for visualization purposes.
    }
    \label{fig:all_task}
\end{figure*}

\section{Experiment}
\subsection{Experiment Setup.}
We design a comprehensive set of experiments to evaluate the performance of our framework, focusing on two key research questions:
(1) Whether the distilled policy derived from our framework enables gentle manipulation;
(2) The stability and robustness of the manipulation strategies produced by our framework;

\subsubsection{Robot system Setup.}
Our experimental platform consists of a UR5 robotic arm equipped with two Gelsight Mini tactile sensors and a parallel gripper. We custom-designed adapters to integrate the tactile sensors onto the parallel gripper, enabling comprehensive acquisition of contact information between the gripper and manipulated objects. Additionally, an Intel RealSense D415 RGB-D camera provides visual perception.
We constructed the corresponding robotic manipulation system within the Isaac Gym simulation environment and utilized the TacSL tactile sensor model to obtain tactile point clouds. This setup minimizes the discrepancy between simulated and real tactile feedback, thereby facilitating the deployment of learned policies in real-world scenarios.
\subsubsection{Benchmark Setup.}
In the simulation environment, we design four manipulation tasks representing diverse challenges. Below is a brief introduction to these tasks:
\begin{itemize}
\item \textbf{Object Stack (OS)}: Generate a random block and a platform. Grasp the block and stack it onto the platform.
\item \textbf{Open and Place (OP)}: Grasp the handle of a box and open it. Move to an object and place the object inside the box.
\item \textbf{Cabinet and Place (CP)}: Move the robot arm to the drawer and open it. Then, move to the object, grasp it, and place it inside the drawer.
\item \textbf{Pour Ball (PB)}: Grasp a cup containing a small ball, move it above a bowl, and rotate the cup to pour the ball into the bowl.
\end{itemize}
As illustrated in Fig.~\ref{fig:all_task}, these long-horizon tasks encompass a variety of fundamental skills required in real-world scenarios. Notably, all tasks leverage the same atomic skill library, demonstrating the scalability and adaptability of our framework across diverse manipulation tasks. 

\subsubsection{Training Setup and Model Preparation}
We record the success rates of each atomic skill as shown in Tab.~\ref{tab:table1}. All skills demonstrate high performance due to their single-stage contact manipulation characteristics. This provides a strong guarantee for generating long-horizon gentle manipulation data.
\begin{table}[h]
    \centering\small
    \begin{tabular}{ccccccc}
    \toprule
    & Grasp & Rotate & Move  & Lateral move & Horizontal pull \\ 
    \hline
    &0.98 & 0.96 & 0.89 & 0.92 & 0.95 \\
    \bottomrule
    \end{tabular}
    \caption{Success Rates of Gentle Atomic Skills Trained via Reinforcement Learning in Simulation.}
    \label{tab:table1}
\end{table}
Our VT-DP utilizes a Transformer-based noise prediction network with approximately 13.27 million parameters. Training involves 100 denoising steps to progressively reconstruct action sequences from noisy expert trajectories, while inference employs a reduced 10-step denoising schedule to balance efficiency and performance.
The policy operates on an 8-step horizon, consuming 2-step observations to predict 5-step future actions. Inference runs at 28 Hz on an RTX 3060 GPU, executing two actions at each step to ensure real-time responsiveness and task success. Training converges after approximately 800 epochs.

\begin{table}[htp]
    \centering\small
    \begin{tabular}{cccccccc}
    \toprule
    VLM Model & OS & OP &CP &PB &Avg \\ 
    \hline
    Doubao-pro & \textbf{0.96} & \textbf{0.81} &0.5 & \textbf{0.58}  & \textbf{0.713}\\
    Doubao-lite & 0.19 & 0.20 &0.22 & 0.13  &0.185\\
    Qwen-max  & \textbf{0.96} & 0.55 & \textbf{0.67}  & 0.50  & 0.670\\
    Qwen2.5-7b & 0.13  & 0.00 &0.00 & 0.00  & 0.033\\ 
    Qwen2.5-32b & 0.86  & 0.65 &0.56 & 0.28  & 0.588\\ 
    Qwen2.5-72b & 0.91  & 0.54 &0.51 & 0.47  & 0.608\\ 
    \bottomrule
    \end{tabular}
    \caption{Success Rates of Trajectory Generation for Different VLM Models and atomic Skill Compositions Across Multiple Tasks.}
    \label{tab:table2}

\end{table}

\subsection{Ablations}
\subsubsection{VLM-Planner Ablation}
We investigated the combination of different VLMs with atomic skills to demonstrate the generalizability of our framework. For each VLM, we utilized pretrained weights directly without any task-specific fine-tuning. The system prompt, environment RGB images, and relevant task descriptions were kept consistent across experiments. Our evaluation focuses on VLM accuracy in atomic skill planning, and transition point judgment. We recorded the success rates of trajectory generation across multiple tasks, with the experimental results summarized in Tab.~\ref{tab:table2}.
Among the tested VLMs, the Qwen2.5-vl-7b-instruct model exhibited significantly lower success rates, likely due to its smaller model size and limited instruction-following capability, which hindered accurate planning and skill transitions. In contrast, Doubao-vision-pro demonstrated superior performance in both atomic skill planning and transition point detection. Considering its strong performance and ease of use, we adopt the Doubao-vision-pro model combined with atomic policies to generate expert demonstrations within the VASK framework.
\begin{table}[h]
    \centering\small
    \begin{tabular}{ccccc}
    \toprule
    Method & OS & OP &CP &PB\\ 
    \hline
    \toprule
    &     \multicolumn{4}{c}{\textbf{Success Rate}} \\ 
    \hline
    VASK+VT-DP (ours) & \textbf{0.90} &\textbf{0.73} & \textbf{0.63}  & \textbf{0.90}\\
    VLM Planning & 0.63  & 0.50 & 0.00 & 0.37\\
    Human data+VT-DP & 0.30  & 0.47 &0.27 & 0.60\\
    \bottomrule
    
    \toprule
    &   \multicolumn{4}{c}{\textbf{Average Success Path Length}}\\
    \hline
    VASK+VT-DP (ours)  &\textbf{85} &\textbf{184} &\textbf{179} &\textbf{175}\\
    VLM Planning  &170 &306 &$/$ &345\\
    Human data+VT-DP  &304 &648 &763 &582\\
    \bottomrule

    \toprule
    &   \multicolumn{4}{c}{\textbf{Average Contact Force (N)}}\\
    \hline
    VASK+VT-DP (ours)  &\textbf{0.322} &\textbf{0.170} &\textbf{0.489} &\textbf{0.091}\\
    VLM Planning  &0.672 &0.482 &$/$ &0.205\\
    Human data+VT-DP  &0.809 &0.493 &0.932 &0.196\\
    \bottomrule
    
    \end{tabular}
    \caption{Simulation Benchmarking Against Extended Baselines: Comparative Evaluation of Success Rate (SR), Average Success Path Length (SPL), and Average Contact Force (ACF) Across Our Method, VLM Planning, and Human Demonstration-Based Approaches.}
    \label{tab:table4}
\end{table}
\subsubsection{Skill Ablation}
We further investigated the direct use of pretrained VLM to generate end-effector waypoints and binary gripper commands by leveraging a knowledge base, manual task descriptions, and background images, with the goal of reducing dependence on atomic skills. As shown in Tab.~\ref{tab:table4}, this approach exhibits relatively low success rates, mainly due to the inherent instability of the VLMs and their inability to effectively handle physical interactions and collisions between the robotic arm and objects. This limitation is particularly evident in complex tasks; for example, the CP task requires large-angle reorientation after opening a drawer, which poses significant challenges for VLM-based waypoint planning. Moreover, this method fails to adequately incorporate contact force information into the manipulation process, rendering it incapable of achieving gentle interactions.

\subsubsection{Force-Aware component Ablation}
\begin{figure}[htp]
    \centering
    \includegraphics[width=0.8\linewidth]{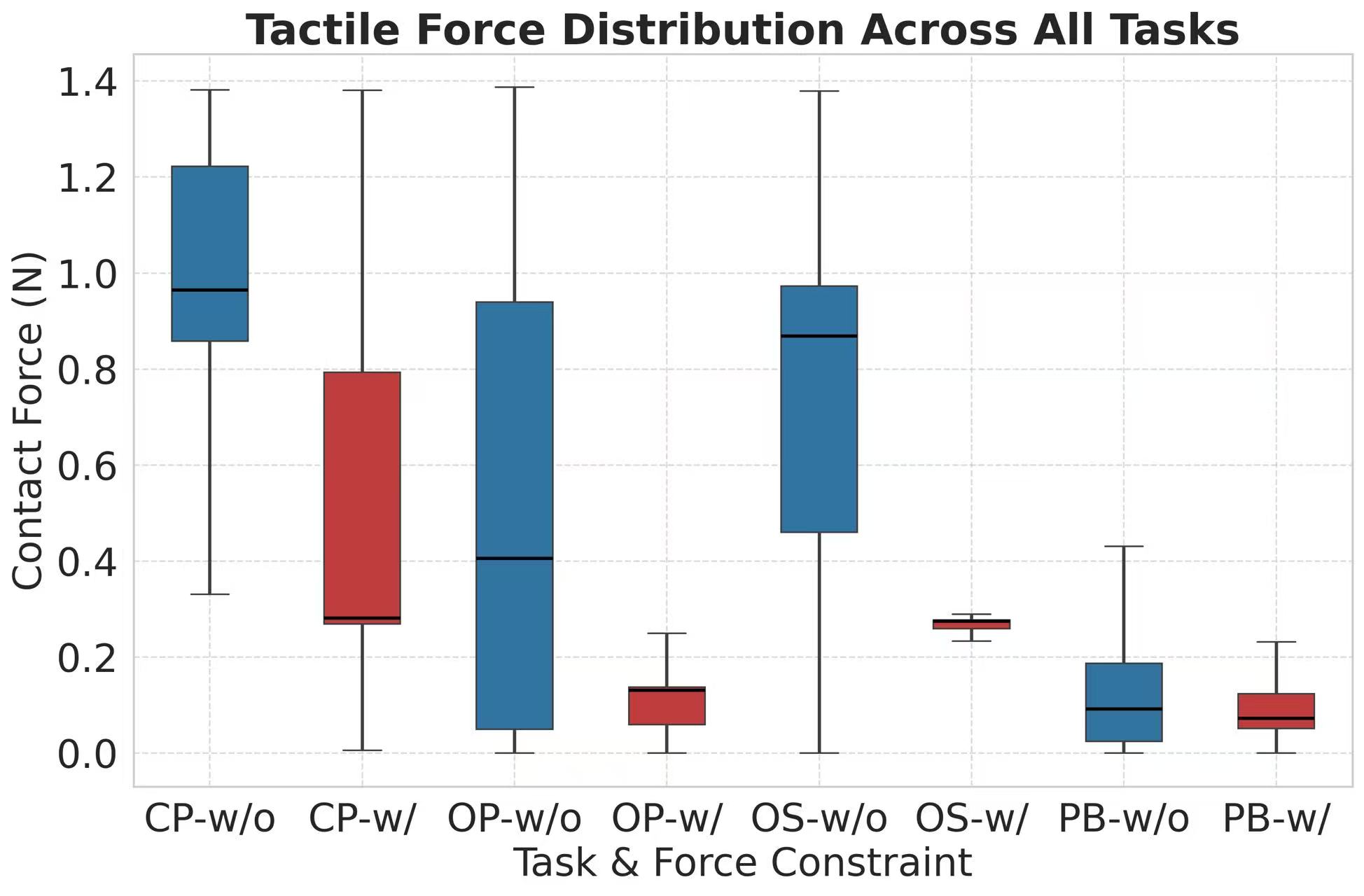}
    \caption{Contact force distribution over ten trials for each of the four manipulation tasks. Blue boxplots show policies trained without force-related penalties, while red boxplots show policies trained with force-aware rewards.
    }
    \label{fig:force}
\end{figure}

To validate the effectiveness of incorporating force-aware rewards for gentle manipulation, we conduct ablation experiment comparing policies trained with and without explicit force-related penalties in the reward function. During long-horizon executions, contact forces from bilateral tactile sensors were continuously measured, excluding non-contact periods. Fig.~\ref{fig:force} illustrates the distribution of contact forces during long-horizon manipulation tasks executed by two different policies. The results indicate that the policy trained with force-aware rewards consistently maintains lower and more stable contact forces, thereby validating that explicitly incorporating contact force factors effectively promotes gentle manipulation.
\vspace{-0.2cm}
\subsection{Method Comparisons}
\begin{table*}[h]
    \centering\small
    \begin{tabular}{cccccccc}
    \toprule
    Method & OS(SR) & OP(SR) &CP(SR) &PB(SR) & Freq(Hz) &Avg(SR) &Task Var\\ 
    \hline
    PC(Vision w/ Tactile) & \textbf{0.90} & \textbf{0.73} &\textbf{0.63} & \textbf{0.90} &28 & \textbf{0.79} &\textbf{0.0134}\\
    PC(Vision w/o Tactile) & 0.83 & 0.30 &0.33 & 0.86 &28 &0.58 &0.0705\\
    RGB(Vision w/ Tactile) & 0.50  & 0.20 &0.00 & 0.43 &9 & 0.28 &0.0389 \\ 
    RGB(Vision w/o Tactile) & 0.43 & 0.23 & 0.00  & 0.37 &8 & 0.26 &0.0274\\    
    \bottomrule
    \end{tabular}
    \caption{Success rates of diffusion-based manipulation policies using different perception modalities as observations. For fairness, all methods utilize the same number of expert demonstrations and identical data generation models.}
    \label{tab:table3}
    \vspace{-0.2cm}
\end{table*}

\subsubsection{Demonstration Data Comparison}
In simulation, we constructed expert demonstration data by manually controlling the robotic arm to complete long-horizon manipulation tasks and subsequently distilled the data using the VT-DP framework. However, we observed that models trained with this approach exhibited suboptimal performance as shown in Tab.~\ref{tab:table4}, frequently suffering from operation stagnation and failure to successfully grasp objects during real-world execution. We attribute these issues to the low quality of manually collected data. Human operation inevitably introduces pauses and erroneous actions during data collection, and the resulting trajectories are subject to subjective biases, often leading to demonstration sequences significantly longer than those generated by our method, resulting in the extracted policies exhibiting higher average path lengths when completing tasks compared to our approach. Moreover, manual data collection incurs substantial labor costs, requiring considerably more time to acquire an equivalent amount of data compared to our framework, while also failing to effectively incorporate contact force information.
In contrast, our framework efficiently acquires robust long-horizon, smooth manipulation policies with reduced data collection effort, achieving superior success rates across diverse tasks. These results demonstrate the effectiveness and practical value of our approach.

\subsubsection{Input Modalities Comparison}

Robot perception involves both RGB images and point clouds, each offering distinct representational advantages. We systematically evaluated diffusion policies trained on these modalities, conducting 30 trials per task, with all demonstration data generated through the VASK framework to ensure fairness. As shown in Tab.~\ref{tab:table3}, the diffusion policy trained solely on RGB images, represented by the method DP \cite{Chi2023DiffusionPV}, performed poorly across all tasks. Incorporating tactile information brought only marginal improvements, which may be attributed to the inherent difficulty of extracting robust features from RGB data alone, especially in long-horizon manipulation tasks that require precise spatial reasoning. The DP3 method, which utilizes visual point clouds, achieved better results owing to their compact encoding of three-dimensional spatial information. However, it still underperformed on complex, contact-rich tasks such as OP and CP, indicating that spatial data alone cannot fully capture the nuanced tactile interactions these tasks demand. Our proposed approach, \textbf{VT-DP}, which fuses visual and tactile point clouds, consistently outperformed all other methods across tasks. This fusion enhances spatial perception through tactile feedback and improves feature representation, resulting in superior overall task performance.

\subsection{Qualitative Experiments in the Real World}
\begin{figure}[htp]
    \centering
    \includegraphics[width=1.0\linewidth]{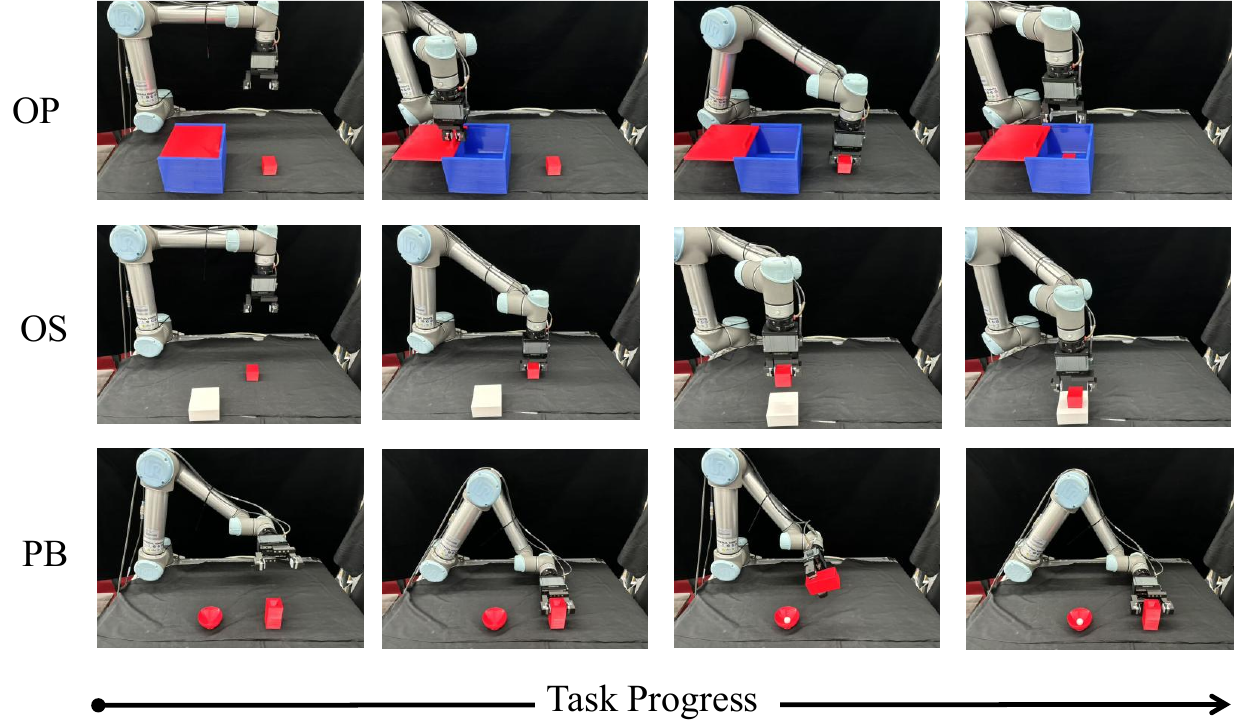}
    \caption{The execution process of long-horizon manipulation tasks in the real world.
    }
    \label{fig:real_task}
    \vspace{-0.2cm}
\end{figure}

We deploy our policy on physical robots using a digital twin system, validating the feasibility of our approach. Specifically, policy observations include proprioceptive data and visuo-tactile point clouds. We apply linear transformations to real-world proprioceptive measurements to align with simulation observations. For visual point clouds, we capture real scenes using D415 cameras; however, these point clouds exhibit quality limitations, particularly in robotic arm regions. To address this, we first transform simulated and real-world point clouds into a unified coordinate system, then merge simulated robotic arm point clouds with real-world scene point clouds. The digital twin ensures state consistency between physical and simulated robotic arms. Additionally, we employ color-based thresholding and point cloud cropping to remove irrelevant environmental backgrounds, achieving representations similar to simulation environments. For tactile point clouds, we align GelSight Mini sensor data with simulated tactile point clouds and perform corresponding downsampling. As shown in Fig.~\ref{fig:real_task}, this method enables deployment of our manipulation policies in real-world environments.

\subsection{Limitations}
Although the current framework significantly reduces the manual effort required to generate demonstration data, the integration of VLM incurs additional computational overhead. Future research will focus on narrowing the sim-to-real gap to facilitate effective deployment in real-world scenarios.

\section{Conclusion}
In this work, we propose a novel framework that decomposes complex manipulation tasks into compositions of atomic skills and leverages visual language models for task planning and decision-making. Combined with a Visual-Tactile Diffusion Policy (VT-DP), the method successfully distills long-horizon gentle manipulation strategies from expert demonstrations. Notably, our approach explicitly incorporates tactile perception into the generation of long-horizon manipulation data, enhancing the capability of the system to perform delicate and contact-rich operations. Since the policy is fully trained in simulation without requiring additional manual demonstration data, the framework significantly reduces both labor and data collection costs. Additionally, the integration of visual language models with atomic skills provides strong scalability, enabling the framework to generalize across a wide range of long-horizon manipulation tasks.

\bibliography{aaai2026}

\end{document}